\documentclass[a4paper, final, 12pt]{article}

\usepackage{eurosym}
\usepackage{amsmath, amssymb, latexsym, amscd, amsthm,amsfonts,amstext}
\usepackage[mathscr]{eucal}
\usepackage{graphicx}
\usepackage{subfig}
\usepackage{float}
\usepackage{color}
\usepackage{hyperref}
\usepackage[utf8]{inputenc}
\usepackage[english]{babel}
\usepackage{algorithm}
\usepackage{algorithmic}
\usepackage{indentfirst}

\newtheorem{definition}{Definition}[section]

\numberwithin{equation}{section}
\begin{document}

\title{Application of Deep Q Learning with Simulation Results for Elevator Optimization}

\author{Zheng Cao$^1$, Raymond Guo$^2$, Caesar M. Tuguinay$^3$, \\
Mark Pock$^4$, Jiayi Gao$^5$, Ziyu Wang$^6$  
\and University of Washington, Seattle, USA 
\and Department of Mathematics
\and zc68@uw.edu$^1$
\and Department of Computer Science \& Math
\and rpg360@uw.edu$^2$
\and Department of Mathematics
\and ctuguina@uw.edu$^3$
\and Department of Computer Science
\and markpock@uw.edu$^4$
\and Academy for Young Scholars
\and jerrygao@uw.edu$^5$
\and Department of Economics
\and ziyuw5@uw.edu$^6$
}
\date{}
\maketitle

\begin{abstract}
This paper presents a methodology for combining programming and mathematics to optimize elevator wait times. Based on simulated user data generated according to the canonical three-peak model of elevator traffic, we first develop a na\"ive model from an intuitive understanding of the logic behind elevators. We take into consideration a general array of features including capacity, acceleration, and maximum wait time thresholds to adequately model realistic circumstances. Using the same evaluation framework, we proceed to develop a Deep Q Learning model in an attempt to match the hard-coded na\"ive approach for elevator control. Throughout the majority of the paper, we work under a Markov Decision Process (MDP) schema, but later explore how the assumption fails to characterize the highly stochastic overall Elevator Group Control System (EGCS).

% These methods are all describing the same thing

\end{abstract}

\vspace{0.5em}


\newpage

\tableofcontents

\newpage

\section{Introduction}
Elevators figure strongly into the daily life of the ordinary urbanite. The role of elevator wait times may seem negligible, but this comes as a result of decades of optimizations and improvements in EGCS facilitated by a wide array of fields. The minimization of elevator wait times becomes especially crucial during the down-peak and up-peak, when many elevators are crowded. Poor algorithmic design results in frustrated and tired workers cramming around elevator doors, wasting valuable time.

This research group has approached elevator optimization via two angles, addressed by two separate teams -- explicit mathematical modelling and machine learning for approximate optimization. This paper focuses on the latter approach. Our team's source code is contained in the ``Elevator Project" GitHub. \cite{EP} We first generate data according to the canonical three-peak model, and use it to build a base-case model to analyze the performance of traditional elevator design. We subsequently turn towards Deep Q Learning to attempt optimization over the na\"ive base-case.

\subsection{Framing and Literature}
We frame our discussion of elevator optimization through the hierarchical paradigm of Elevator Group Control Systems (EGCS), the central mechanisms in multi-elevator buildings which control and monitor elevator motion. Where elevators stay by default, which elevators will be dispatched to various hall calls, etc. are managed by EGCS. Its importance to internal transportation has led to an array of research where innovations from across engineering disciplines have been combined and synthesized to produce the modern elevator -- a far cry from the early days of elevators which ran on schedules. Throughout the paper, we interpret the idea of EGCS in a way abstracted from its material implementations -- as the overall state, algorithm for responding to that state, and state transition functions in a particular building.

Several authors before us have attempted to apply machine learning to EGCS -- particle swarm optimizations \cite{Bolat}, Convolutional Neural Networks (CNNs), and neuro-fuzzy systems \cite{neurofuzzy}, amongst other approaches. Combining machine learning approaches with more rigorous mathematical approaches has been a particularly fruitful approach to other problems -- for example, Zheng Cao's previous paper ``Application of Convolutional Neural Networks with Quasi-Reversibility Method Results for Option Forecasting." \cite{Z} 

 \subsection{Strategy}
Our approach to optimization is straightforward -- an application of Deep Q Learning to what we characterize as a classification problem for an arbitrary decision algorithm taking a simplified version of a building's current state and outputting commands to the elevator(s) therein.

\section{Theoretical Background}

\subsection{Reinforcement Learning, Markov Decision Processes}
The generic Reinforcement Learning (RL) problem is framed as an interaction between an agent and an environment. At each time-step, the agent selects an action out of a set of possibilities. The environment responds by shifting to a different state and presenting that state to the agent alongside a scalar reward. This interaction continues until the environment reaches a terminal state (there are also RL problems involving environments without a terminal state, but the problem discussed here is not one of them). A complete sequence of actions from the agent and responses from the environment, from start to terminal state, is known as an ``episode". We denote the $n$th state by $S_n$, the $n$th action by $A_n$, and the $n$th reward by $R_n$ (where $S_0$ is the initial state, $A_0$ is the first action, and $R_0$ is the reward given in response to that action).

In this sense, RL problems can be thought of as a series of classification problems where the agent, at every time step $n$, is tasked with choosing the action that maximizes some function of the rewards following the $n$th action. This function is usually (and is here),
\[\sum_{i=n}^t \lambda^{i-n}S_i\]
known as the discounted return, where $t$ is the time step after which the terminal state is reached. $0 \leq \lambda < 1$ is the ``discount factor" and chosen as a hyperparameter in training, where lower values will make the model prioritize increasing immediate rewards, and higher values will make the model have a more ``long term" value.

A finite Markov Decision Process (MDP) is a special case of these RL problems where the number of states and possible rewards is finite, the number of actions that can be chosen in response to each state is finite, and most importantly, the probability of any state-reward pair given in response to any action and previous state is dependent only on that action and previous state (and not any of the actions and states that preceded them). This is known as the ``Markov Property" and can be expressed symbolically by
\[\text{Pr}\{R_{t+1} = r, S_{t+1} = s | S_0, A_0, R_1, . . . , S_{t-1}, A_{t - 1}, R_t, S_t, A_t\} = \text{Pr}\{R_{t+1} = r, S_{t+1} = s | S_t, A_t\}\]
for any $r,s$ that lie in the set of possible rewards and states respectively. A more formal definition is given in the appendix. MDPs are important because most proofs that provide guarantees that RL methods converge are based on the MDP case, although empirically great success has been achieved in applying these methods to non-MDP RL problems. Its importance will be expounded upon in a later section. \cite{RLAI}

\subsection{Q-Learning}

Finite MDPs are solved by finding a good policy $\pi(a|s)$, which is a probabilistic function that defines the actions of the agent. In particular, $\pi(a|s)$ denotes the probability of the agent choosing action $a$ next, if the last state were to be $s$. With respect to such a policy $\pi$, we can define a value function $v_\pi(s)$, which defines the expected return of an agent starting at state $s$ and following policy $\pi$. Similarly, we can define an action-value function $q_\pi(s,a)$ which defines the expected return of an agent starting at state $s$, taking action $a$, and then following policy $\pi$ thereafter. In finite MDPs, there is at least one optimal policy \cite{RLAI} $\pi_*$, whose value function $v_*$ has the property that for any other value function $\pi$, $v_*(s) \geq v_\pi(s)$ for all states $s$. Its action-value function is denoted $q_*$. The optimal action-value function must satisfy the Bellman equations \cite{RLAI}, shown below:
\[\forall \text{ states }s, q_*(s,a) \mathbb{E}\left[R_{0}+\gamma \max_{a'}q_*(S_{1},a') | S_0=s,A_0=a \right]\]

In fact, the optimal action-value function is the only action-value function capable of satisfying this. \cite{RLAI}

It essentially states that the action-value function must be the expected value of the first reward provided added to the action-value function applied to the next state and best action given that state.

Many RL methods try to approximate solutions to this equation through alternating policy evaluation (computing the action-value function of the policy you currently have) and policy improvement (actually making the policy better).

Early versions of Monte-Carlo methods performed policy evaluation by computing a collection of episodes and approximating the action-value function at state $s$ and action $a$ by the average of the return from all episodes passing through $s$ using $a$, and performed policy improvement by changing the policy to pick $\pi(s)$ to be the argmax of $q(s,a)$ over all possible actions $a$, where $q$ is the approximate action-value function found in policy evaluation.

These methods have the problem of only updating predictions after (possibly extremely long) episodes complete, so they are improved upon by Q-Learning \cite{RLAI}, which instead, in policy evaluation, uses the update rule
\[Q(S_t,A_t) \longleftarrow Q(S_t,A_t) + \alpha[R_{t+1} + \lambda\max_{a}Q(S_t,a)-Q(S_t,A_t)]\]

where $R_{t+1}$ is the actual reward given by the environment, $Q$ is our current estimate for the action-value function, and $\alpha$ is a hyperparameter which decides how heavily new data should be weighted. It has been proven that Q-learning on an MDP converges almost surely to the optimal value function. \cite{QLearning} This method is still impractical because they require storing estimates for $Q(s,a)$ for every state-action pair $s$, $a$, or at least the ones that are often seen in practice. In our case, and many others, there are simply far too many state-action pairs for this to be done.

As a result, we instead rely on a method first introduced in a paper using RL to play Atari \cite{PADRL} which utilizes a Neural Network to evaluate $Q$. The state, actions, reward, and next state from each time-step are stored in a queue, and every time a finite number of time-steps have occurred, the Neural Network is trained on data in the queue with loss function 
\[Q(S_n,A_n) - (r_n + \max_{a'}Q(S_{n+1},a'))\]
where $Q$ is the current action-value function estimate, $S_n$ is the current state, $A_n$ is the current action, $r_n$ is the reward given for that action, and $S_{n+1}$ is the next state. We note that $(r_n + \max_{a'}Q(S_{n+1},a'))$ is the update value from Q-learning with $\alpha$ set to 1. The policy we follow is dictated by a second neural network, and policy improvement is performed by updating the second neural network's weights with a weighted average of weights from both networks, essentially ``nudging" the policy in the right direction. More details are shown below.

\section{Simulation}
Before tackling EGCS, we seek to develop a system for generation of user traffic data to feed into the system. We modularize the generation into individual people who select times to make hall calls from a truncated Poisson distribution.

We chose to simulate data rather than directly observe it in the real world in order to have the flexibility to work with an arbitrary amount of people while still accounting for the three-peak model. Moreover, we are able to adjust the characteristics of the building with which we are working rather than being bound to observed characteristics. We also save a considerable amount of time and energy.

\subsection{Individual Person Values}

To have the widest application of our model, we generated raw data based on the standard 9-to-5 work schedule with a 30 minute lunch break in between. We assumed that all the workers would arrive and leave the building with a mean time of 32400 seconds (9AM) and 61200 seconds (5PM) with a standard deviation of 1800 seconds (30 minutes). To simulate real life, we also randomly generated each person’s weight based on a normal distribution based off of an individual's sex.

\subsection{Office Building Values}
% \begin{enumerate}
% \item 
% \href{https://buildingandinteriors.com/find-the-magic-number-elevators-per-building/}{Standard num elevator for office}

% \item 
% \href{https://www.eia.gov/todayinenergy/detail.php?id=21152}{Average office building size} and
% \href{https://urbanland.uli.org/development-business/pillars-of-design/}{this}

% \item 
% \href{https://www.rentcafe.com/blog/apartmentliving/high-mid-rise-residential-buildings-overshadowing-low-rise/}{Number of floors in each building}

% \item 
% \href{https://aquilacommercial.com/learning-center/how-much-office-space-need-calculator-per-person/}{Average Space Per Person}
% \href{https://library.density.io/insights/the-most-important-metric-in-corporate-real-estate/}{More}
% \href{https://iofficecorp.com/blog/office-space-per-employee}{and finally}
% \end{enumerate}
Similarly to individual person values, we simulated office building values for simulation; the following describe our office building values and are based off of aggregate personal observation:
There are 8 floors in the building.
There are approximately 200 workers per elevator in each building.
It takes approximately 15 seconds for an elevator to open, load people, and close.
It takes approximately 5 seconds for an elevator to ascend or descend a single floor.

\subsection{Simulated Table}
\textbf{Table 1. Sample Simulated User Data}
\label{elevatordatatab1}
\begin{center}
\begin{tabular}{|l|l|l|l|}
\hline
time (second) & start floor & destination floor) & weight \\ \hline
... & ... & ... & ... \\ \hline
27000 & 1 & 2 & 77.9... \\ \hline
... & ... & ... & ... \\ \hline
27042.9... & 1 & 7 & 78.3... \\ \hline
... & ... & ... & ... \\ \hline
60754.1... & 8 & 1 & 101.6... \\ \hline
... & ... & ... & ... \\ \hline

\end{tabular}
\end{center}

\section{Modeling}

\subsection{Elevator}
We formulate a basic representation of a single-elevator control system which will serve as the later basis for our expansions to more complex EGCS. 

Using one elevator is both the easiest to mathematically model and the easiest to attempt to bring out an optimal method. Therefore, we decided to use a single elevator to evaluate a base-line value.

When there are more elevators, say $n$ elevators, a very simple threshold to compare our results with would simply be to see if our models can improve the time such that it is better than base-line / n seconds, which would be the result if we were to distribute the people evenly across all elevators.

\subsection{Environment}
The environment consists of an object called a Time List containing all the events, an object called a State which describes the certain aspects of the simulation.

In particular, the State consists of the upward and downward calls of passengers looking to enter the elevator and go a specific direction, the time of the simulation, the elevator's speed, the wait time that it takes to open and close the door to let passengers in, and the elevator.

The environment details will be expounded upon in the following na\"ive models we will detail.

\subsection{Model Interactions}
Attached to the environment is the model which decides an action based off of each State. Each action will update the State and Time List accordingly; however, the State may be updated prior to each action based off of events within the Time List.

At the start of the simulation, we initialize the state so that the elevator starts at the first floor and the state's time starts at $27000$ seconds (7:30 AM). During the simulation, we iterate through the Time List until there are no more events within the Time List and when there are no more passengers within the State that wish to be moved. When this occurs, the simulation ends and we calculate the total time all it took for all the passengers to reach their destinations.

The model takes actions based off of the State; however, it only gets passed 3 objects from the State:

\begin{quote}
1. The state of the up / down button of each floor -- Whether it is pressed or not.
\end{quote}

\begin{quote}
2. The state of the buttons in the elevator -- whether it is pressed or not.
\end{quote}

\begin{quote}
3. The current floor of the elevator.
\end{quote}

\subsection{Design}
Our system centers around the management of a global state variable by a main event loop which processes events in a Time List, an ordered procedure for all the relevant events in the period over which the simulation occurs -- for us a day -- generated beforehand.

To be specific, we determine before the start of a simulation when people will enter and exit the building and which floors they will want to go to using the classical three-peak model and truncated Poisson distributions around each peak. The various hall calls thereby generated are fed into our Time List to be processed by the event loop as the simulation progresses.

Treating each Time List event as a stimulus, we manage the current State and request an action from the model, passing in our State as a parameter, to precipitate the succeeding state. Any model we use for this single-elevator system has a simple choice between three commands -- Idle, Move, and Open Doors. This classification problem forms the premise of the optimization we attempt to achieve. Our na\"ive model deals with State in a hard coded and 'common sense' way -- one which is of course deeply inapplicable to more complex EGCS, but allows us to verify the functionality of the system.

See Na\"ive Model with Environment 1.0.

\subsection{Na\"ive Approach to Elevator Control:}
We created a basic algorithm for elevator control that approximates how many elevators practically function. The algorithm alternates between two phases. In phase 1, there are no passengers in the elevator, and the elevator travels to the closest floor with a hall call (if there are no hall calls at all, the elevator will instead idle). Once the elevator travels to a floor with a hall call, it will randomly choose one of the (at most two) directions in which hall calls have been made on that floor, open its doors, and signal its intent to travel in that direction. 

This begins phase 2, where the elevator will continuously travel in that direction, letting people off at their intended destination and picking people up when it reaches a floor with hall calls in the same direction it is travelling in. This phase continues until there are no remaining passengers on the elevator, in which case phase 1 will start again. We tested the results of this model on both the 1.0 (see image Na\"ive Model with Environment 1.0) and 2.0 (see image Na\"ive Model with Environment 2.0) versions of the na\"ive model.

\section{Modeling via Deep Q Network}
To further optimize total run time, we set a Deep Q Networks (DQN) to be the EGCS's inner decision making model.

\subsection{Motivations}
A DQN model was chosen to be the machine learning model to optimize the environment because there exist numerous examples of successful DQN decision making models trained and tested in similar environments to our EGCS. 

The original DQN Atari Paper [4] has shown that DQNs can be trained on environments with large continuous State Spaces and discrete Action Spaces. The EGCS State Space is mildly large and continuous, and the Action Space is discrete.

\newpage

\subsection{EGCS Data Encoding and Decoding}
The following describe how data flows in and out of the inner decision making DQN model within the EGCS.

\begin{algorithm}
\caption*{Input Encoding}
\begin{algorithmic}
\STATE Get capacity, current weight, current position, up buttons, down buttons, buttons pressed from State. 
\STATE tensor = $[\text{cacapacity)}, \text{float(current weight)}, \text{float(current position)}]$

\FOR{entry in buttons pressed (8 entries)}
    \STATE append float(entry) to tensor
\ENDFOR

\FOR{entry in up buttons (8 entries) }
    \STATE append float(entry) to tensor
\ENDFOR

\FOR{entry in down buttons (8 entries)}
    \STATE append float(entry) to tensor
\ENDFOR

\STATE (Note: float(boolean) $== 0$ for false and float(boolean) $== 1$ for true)

\end{algorithmic}
\end{algorithm}

\begin{algorithm}
\caption*{Output Decoding}
\begin{algorithmic}
\STATE Get Action Value: Natural Number between 0 and 4.

\IF{Action Value == 0}
\STATE Action = Idle
\ELSIF{Action Value == 1}
\STATE Action = Open Close Doors for Up Calls
\ELSIF{Action Value == 2}
\STATE Action = Open Close Doors for Down Calls
\ELSIF{Action Value == 3}
\STATE Action = Move Up
\ELSIF{Action Value == 4}
\STATE Action = Move Down
\ENDIF

\end{algorithmic}
\end{algorithm}

\newpage

\subsection{DQN Action Model}
The following describes how the EGCS DQN Action Model is trained.

\begin{algorithm}[!h]
\caption*{Training the EGCS Action Taking DQN}

\begin{algorithmic}[!h]

\STATE Set $\epsilon$ and $\lambda$ values: Number between 0 and 1.
\STATE Initialize Sample Hall Calls (SHC).
\STATE Initialize Updating $Q^{*}$ Steps as $C$.
\STATE Initialize Replay Memory as D.
\STATE Initialize Action Value Model $Q$ with weights $W$.
\STATE Initialize Target Action Value Model $Q^{*}$ with weights $W^{*} = W$.
\STATE Set epoch range as RANGE.

\FOR{Epoch in RANGE}

\STATE Initialize Time List and Initialize State $S_t$ with SHC.
\STATE Total Time = 0, Number Steps = 0.
\WHILE{no more events in Time List}
\IF{action can be taken}
\IF{$\epsilon > $ random number between 0 and 1}
\STATE $A_t$ = Choose Random Action.
\ELSE
\STATE $A_t = \text{ArgMax}_A Q(S_t, A, W)$.
\ENDIF
\STATE Take action $A_t$ and observe $Q_{\text{Val}}$ = $Q(S_t, A, W)$, and $S_{t+1}$.
\ENDIF

\STATE Update Total Time, Number Steps, and $S_{t}$.

\STATE $R_t$ based off of environment: $R_{t} = -1 * $ number of people waiting * sum added time.
\STATE Store transition $(S_t, A_t, R_t, S_{t+1})$ in D.
	
\STATE Sample Random Minibatch of Transitions $(S_j, A_j, R_j, S_{j+1})$ from D as Mini.
\STATE Minibatch Count $= 0$.

\FOR{$(S_j, A_j, R_j, S_{j+1})$ in Mini}

\STATE $M_{j} = \lambda * \text{max}_{A} Q^{*}(S_{j+1}, A, W^{*})$ or $0$ if $S_{j + 1}$ is terminal. Then set $Y_{j} = R_{j} - M_{j}$.

\STATE Loss = $(Y_j - Q(S_j, A_j, W))^2$.

Perform a Gradient Descent Step with Loss Value, with respect to $Q(W)$.
Minibatch Count $+= 1$.

\IF{Minibatch Count \% $C == 0$}
\STATE $Q^{*}(W^{*}) = Q(W)$.
\ENDIF

\ENDFOR
\ENDWHILE
\ENDFOR
\end{algorithmic}
\end{algorithm}

\newpage

The following describes how the EGCS DQN Action Model is used for inference.

\begin{algorithm}
\caption*{Inference with the EGCS Action Taking DQN}

\begin{algorithmic}

\STATE Initialize Sample Hall Calls (SHC).
\STATE Initialize Action Value Model $Q^{*}$ with trained weights $W^{*}$.

\STATE Initialize Time List and Initialize State $S_t$ with SHC.
\STATE Total Time = 0, Number Steps = 0.
\WHILE{no more events in Time List}

\IF{action can be taken}
\STATE $A_t = \text{ArgMax}_A Q(S_t, A, W)$.

\STATE Take action $A_t$ and observe $Q_{\text{Val}}$ = $Q(S_t, A, W)$, and $S_{t+1}$.
\ENDIF

\STATE Update Total Time, Number Steps, and $S_{t}$.

\ENDWHILE
\end{algorithmic}
\end{algorithm}

\section{Results}

\subsection{Model Results}

\begin{figure}[!htb]
\begin{center}
\hspace{20pt}
\includegraphics[scale=0.25]{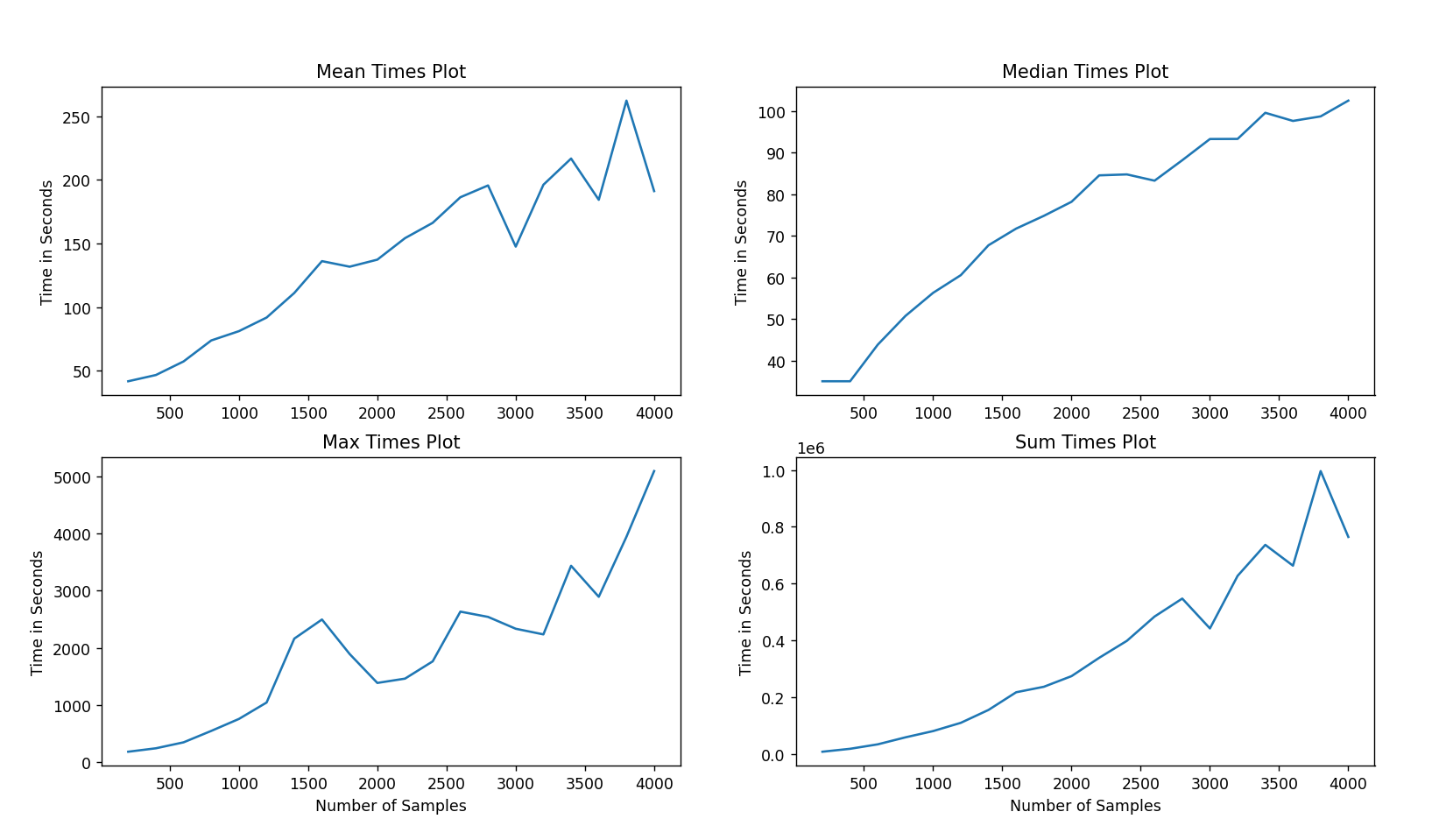}\\
\caption*{Na\"ive Model with Environment 1.0}
\end{center}
\end{figure}

\newpage

\begin{figure}[!htb]
\begin{center}
\hspace{20pt}
\includegraphics[scale=0.25]{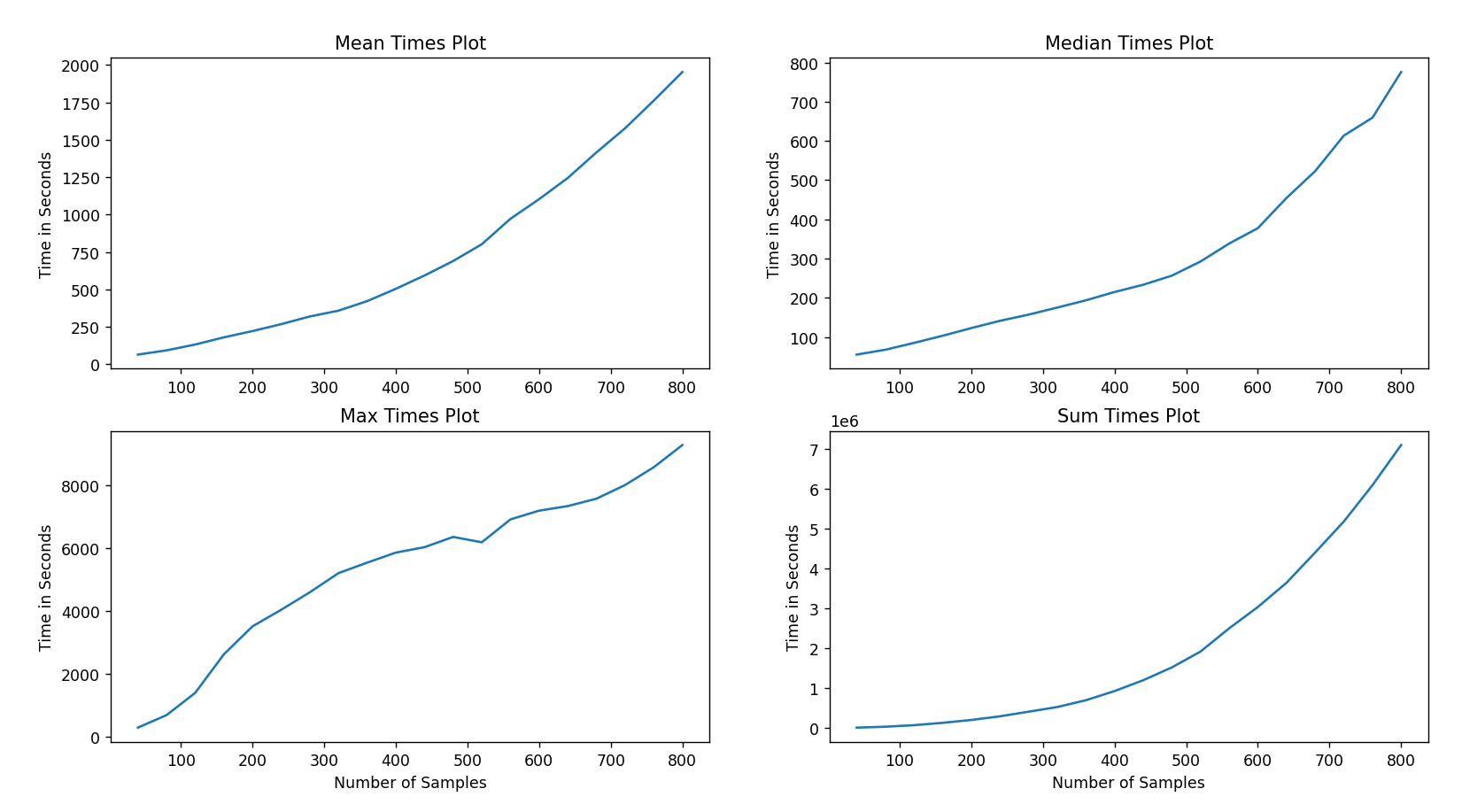}\\
\caption*{Na\"ive Model with Environment 2.0}
\end{center}
\end{figure}

\begin{figure}[!htb]
\begin{center}
\includegraphics[height=105mm, width=60mm]{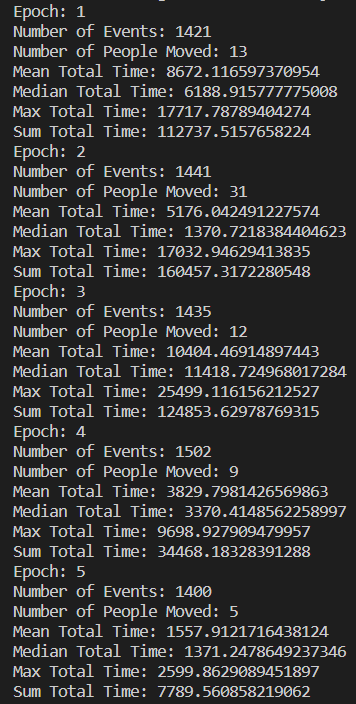}
\includegraphics[height=105mm, width=60mm]{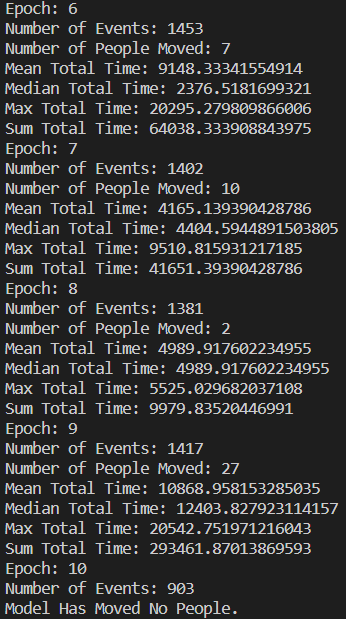}
\caption*{Output logs displaying the training of the DQN Model.}
\label{img12}
\end{center}
\end{figure}

\subsection{Model Failure and Why The Model Failed}

As seen by the above output logs, there is no improvement in Number of Events, Number of People Moved, Mean Total Time, Median Total Time, Max Total Time, Sum Total Time, when examining the  model over ten epochs/full simulation cycles; almost all improvements were abrupt and did not follow a trend of improvement.

The EGCS we are working with is not an MDP. Observe the following: Take any state $s$ such that the immediate state after, $s^{'}$, contains a Hall Call not previously introduced into the EGCS. The Hall Call is placed within the EGCS and the state $s$ is updated to $s^{'}$ at random with no effect from action $a$. Then, there does not exist a function $P_{a}$ for $s, s^{'}, a$ as described by Definition 9.3. Therefore, the EGCS is not an MDP.

We know that such states $s, s^{'}$ exists by examining the simulation at the beginning of time: Time $27000$ seconds (7:30 AM). Progress through the simulation by taking a set of sequential actions $A = \{a_{1},...,a_{n}\}$ with each action being taken on a set of sequential states $S = \{s_{1},...,s_{n}\}$ until we reach a single Hall Call $h$ and no $s \in S$ contains $h$. The Hall Call will then be placed in the next state: $s^{'}$.

An MDP is necessary for any analytic assurance of convergence within Q-learning. \cite{QLearning} DQN is an implementation of Q-learning where we approximate the reward function with a neural network. Extrapolating from \cite{QLearning}, there is no assurance for convergence within the DQN model proposed in this paper, from the perspective of the paper from \cite{QLearning}. Additionally, the authors of the original DQN Atari Paper \cite{PADRL} said that their method lacks any convergence guarantees.

\subsection{Possible Solutions}

To fix the problem of having no analytic assurance of convergence, a modified EGCS where the inner decision making model has knowledge on all the future Hall Calls can be created.

The complication with this is that the EGCS no longer models real-life scenarios: Elevators should not know when and how many people will make Hall Calls. Therefore, this model will not be useful outside of simulations.

A possible fix to this modified EGCS inner decision making model previously described is to create a model which will predict future Hall Calls. This will attach to the inner decision making model and will feed it predicted future Hall Calls. It could back-propagate and update its weights based off of the Hall Calls the EGCS actually receives during each run.

\section{Future Work}

The largest piece of future work we would like to see is a more successful RL-based model that offers improvements over the Na\"ive algorithm we created. We would also like to see the environment expanded to handle multi-elevator systems and create a more generalized Na\"ive algorithm that works on these multi-elevator systems.

\section{Conclusion}

In this project, we created a simulated environment to test the effectiveness of a given elevator control algorithm and demonstrated its correctness with a na\"ive control algorithm that achieved expected results. In the future, this environment can be used to test various other Reinforcement Learning approaches to our elevator optimization problem. We also attempted to employ a Deep Q-Learning approach to create a ML model to improve upon the na\"ive algorithm discussed above. This goal will be addressed more completely in future work, and we hope that the environment we created can be used as groundwork for future attempts based on Reinforcement Learning to solve the same problem.

\section{Acknowledgement}

The original inspiration for elevator optimization came from Zheng Cao and his experiences with the antiquated elevators on the University of Washington campus. From this catalyst, we have attempted to approach elevator optimization using reinforcement learning and explicit mathematical modelling. Our second approach, facilitated by our math research team, we are currently outlaying in a separate paper, ``Application of Spatial Process and Gibbs Random Field Approaches for Dumbwaiter Modeling for Elevator Optimization," which applies methods from pure math such as Spatial Process and  Gibbs Random Field approaches to optimize elevator wait times. \cite{ASPGRFADMEO}

Both projects produce independent results, and potential combinations of both approaches may help achieve a better optimization in future research. Nevertheless, we acknowledge the help the main authors of the additional paper have serviced in creating this reinforcement learning approach: Wanchaloem Wunkaew, Xiyah Chang, and Benjamin Davis.

\newpage

\section*{Appendix}

\subsection*{Markov Decision Processes}

A Markov Decision Process (MDP) is defined as a four-tuple $(S,A,P_{a}, R_{a})$ with the following properties:

\begin{definition}
S is a set of states, which is called a State Space.
\end{definition}

\begin{definition}
A is a set of actions, called the Action Space. The Action Space is (usually) derived from S.
\end{definition}

\begin{definition}
$\forall a \in A$ and $\forall s,s^{'} \in S$, $P_{a}(s,s^{'}) = Pr(s_{t+1} = s^{'} | s_{t} = s, a_{t} = a)$; the probability of taking action $a$ at state $s$ and transitioning to state $s^{'}$.
\end{definition}

\begin{definition}
$\forall a \in A$ and $\forall s,s^{'} \in S, R_{a}(s,s^{'}, a)$ is the immediate reward received after taking action $a$ and transitioning from state $s$ to state $s^{'}$.
\end{definition}

\end{document}